**Do AI models produce better weather forecasts than physics-based models? A quantitative evaluation case study of Storm Ciarán**

Andrew J. Charlton-Perez[1] (0000-0001-8179-6220), Helen F. Dacre[1] (0000-0003-4328-9126), Simon Driscoll[1] (0000-0001-5080-1234), Suzanne L. Gray[1] (0000-0001-8658-362X ), Ben Harvey[1,2] (0000-0002-6510-8181), Natalie J. Harvey[1] (0000-0003-0973-5794), Kieran M. R. Hunt[1,2] (0000-0003-1480-3755), Robert W. Lee[1] (0000-0002-1946-5559), Ranjini Swaminathan[1,3] (0000-0001-5853-2673), Remy Vandaele[1,3] (0000-0002-0693-8011), Ambrogio Volonté[1,2] (0000-0003-0278-952X)

[1]Department of Meteorology, Univ. of Reading, Reading, United Kingdom

[2]National Centre for Atmospheric Science, Univ. of Reading, Reading, United Kingdom

[3]National Centre for Earth Observation, Univ. of Reading, Reading, United Kingdom

**Abstract**

There has been huge recent interest in the potential of making operational weather forecasts using machine learning techniques. As they become a part of the weather forecasting toolbox, there is a pressing need to understand how well current machine learning models can simulate high-impact weather events. We compare forecasts of Storm Ciarán, a European windstorm that caused sixteen deaths and extensive damage in Northern Europe, made by machine learning and numerical weather prediction models. The four machine learning models considered (FourCastNet, Pangu-Weather, GraphCast and FourCastNet-v2) produce forecasts that accurately capture the synoptic-scale structure of the cyclone including the position of the cloud head, shape of the warm sector and location of warm conveyor belt jet, and the large-scale dynamical drivers important for the rapid storm development such as the position of the storm relative to the upper-level jet exit. However, their ability to resolve the more detailed structures important for issuing weather warnings is more mixed. All of the machine learning models underestimate the peak amplitude of winds associated with the storm, only some machine learning models resolve the warm core seclusion and none of the machine learning models capture the sharp bent-back warm frontal gradient. Our study shows there is a great deal about the performance and properties of machine learning weather forecasts that can be derived from case studies of high-impact weather events such as Storm Ciarán.

**Introduction**

During the 20th century and the first two decades of the 21st century, numerical weather prediction (NWP) transformed atmospheric science[1]. The combination of physical and mathematical understanding, the availability of high-performance computing and the expansion of the network of Earth system observation led to remarkable and continued progress in the skill and availability of weather forecasts. Numerical weather predictions are a ubiquitous part of modern life, with applications on many different timescales and in sectors as diverse as transport, agriculture, healthcare and recreation.

Over the last two years, machine learning (ML) techniques, a subset of the rapidly developing field of artificial intelligence (AI), have begun to be applied to the weather prediction problem in earnest. Whilst ML has had applications in climate science for many decades[2-4], with these communities aware of its potential[5], and is increasingly used for post-processing weather forecasts[6,7], recent advances in ML and advancements in GPUs (Graphics Processing Units), have enabled the beginning of a 'new dawn' in the application of ML and AI techniques to weather and climate prediction[8].

The publication of the WeatherBench dataset[9] and the 10-year roadmap for ML use by the European Centre for Medium Range Weather Forecasts (ECMWF)[10], amongst other developments, stimulated interest and investment in the development of ML-models for weather forecasting. During 2022 and 2023, four ML-models were developed by major technology companies to address the short to medium-range (0-10 day) forecasting problem. These models have all been shown to produce skillful 0-10 day forecasts of the 500 hPa geopotential height field, based on the widely used Anomaly Correlation Coefficient metric[11]. All four models use an encode-process-decode framework but with differing architectures:

- FourCastNet[12], developed by NVIDIA and based on Fourier Neural Operators (FNO) with a vision transformer architecture;

- FourCastNet version 2[13], which builds on FourCastNet by using spherical FNOs;

- Pangu-Weather[14], developed by Huawei and based on a three-dimensional Earth specific transformer and hierarchical temporal aggregation; and

- GraphCast[15], developed by Google DeepMind and based on graph neural networks.

Similar techniques have been used to develop models for other forecast tasks (e.g. MetNet-3 for 12-hour precipitation forecasts in the contiguous United States and 27 European countries[16]). At the present time, ML-models primarily produce deterministic forecasts, but rapid progress is being made in producing fully probabilistic forecasts[17-19]. All four ML-models are extremely efficient when run on GPU or TPU (tensor processing unit) devices, typically producing 10-day forecasts in a few minutes.

Given the infancy of ML-model weather prediction, to the authors knowledge, there are no prior studies that compare how the four ML-models and NWP models capture individual, impactful weather events. Examination of individual weather events available from the papers which describe the ML-models are limited to qualitative comparisons of the simulation of tropical cyclones and atmospheric rivers by FourCastNet[12] and quantitative assessment of the track error of tropical cyclones by Pangu-Weather[14] and GraphCast[15]. There are no published studies that examine ML-model forecasts of extratropical windstorms[20], despite their potential to cause multi-billion dollar damages[21] and increasing severity under climate and population change[22].

In this study, we therefore seek to advance knowledge of the comparative performance of ML and NWP models by comparing their forecasts of Storm Ciarán, which affected several European countries during November 2023. This is a valuable out-of-sample test for the ML-models because their training datasets all end before the beginning of 2023. We compare the ability of the models to capture the detailed physical structure of the storm and its impacts at two lead times over which operational weather forecasters were actively engaged in issuing weather warnings to the public. An accurate description of the physical structure of this, or any other, storm is a key component of the forecasting its compound impact[23] and in constructing plausible storylines for end-users[24].

**Results**

Storm Ciarán and its associated impacts

Storm Ciarán was first seen as a low-pressure weather system south of Newfoundland at about 00 UTC 31 October 2023. Based on surface analysis charts issued by the UK Met Office, it then tracked quickly across the North Atlantic, undergoing explosive deepening from 988 hPa at 00 UTC on 1 November to 954 hPa at 00 UTC on 2 November at which time it was located to the northeast of France. This deepening rate, 34 hPa in 24 hours means that Ciarán was an extratropical cyclone "bomb"[25]. The lowest pressure recorded, 953 hPa at 06 UTC 2 November, is a record low pressure for a November storm observed in England[26]. Figure 1 shows surface observations of the 10-m wind speed, cloud cover and mean sea level pressure (MSLP). The cyclonic circulation around the storm centre (with the lowest MSLP observed on the English south coast near the Isle of Wight) has a maximum wind speed of 65 knots on the Normandy coast in France.

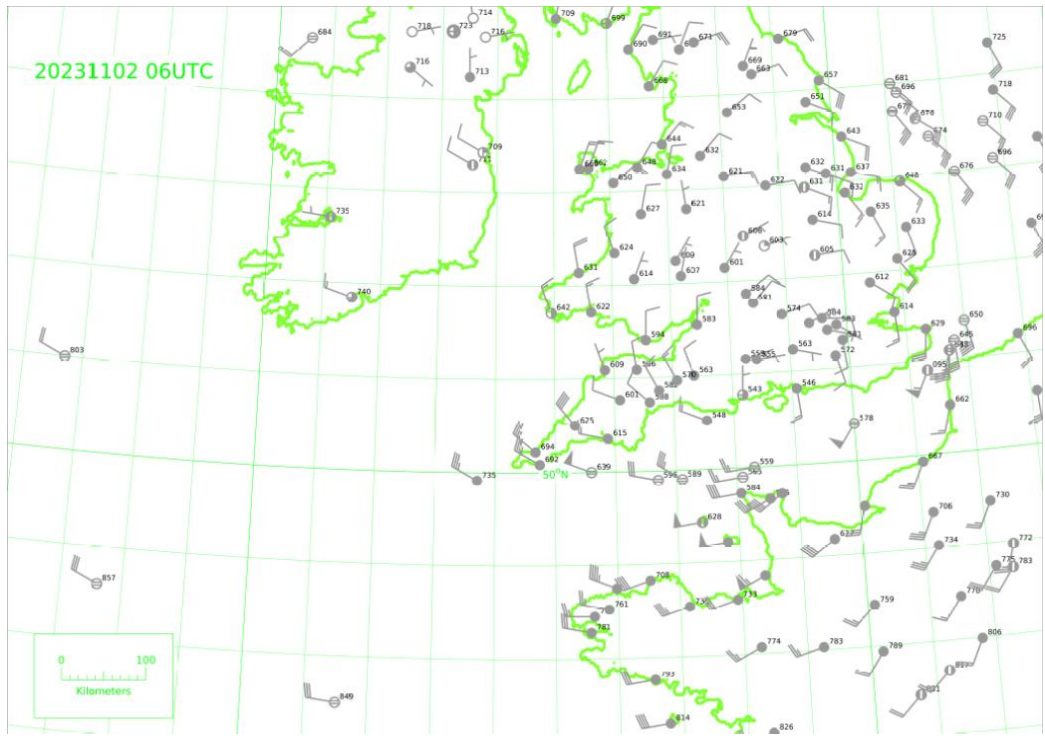

*Figure 1. Surface land and ship station SYNOP observations of Storm Ciarán at 06 UTC 2 November 2023 extracted from the MetDB database[27,28], which holds data including surface and upper air observations and some satellite data. The observations are shown as simplified station circles using conventional notation[29]. Circle shading indicates cloud cover in octas, wind barbs and feathers indicate wind speed in knots with the wind direction towards the circle, and numbers are the last three digits, including a decimal place, of the MSLP (in hPa) e.g., 543 equates to 954.3 hPa. Some thinning of observations has been performed for clarity and note that two ships both reported at 51.1°N, 1.7°E with different wind speeds and directions.*

Although Storm Ciaran was not a classic Shapiro-Keyser cyclone[30], clear banding in vicinity of the tip of the cloud as it encircles the storm centre to the poleward side (called the cloud head) could be seen in satellite imagery before it made landfall in northern France. This banding suggests that a sting jet may have been present in Storm Ciarán[31], however its identification requires methodologies beyond the scope of this study. Gusts of over 100 knots (51 m s$^{-1}$) were reported in several locations in Brittany[32], with a maximum of 111.7 knots (57.5 m s$^{-1}$) recorded at Pointe du Raz at approximately 0200 UTC on 2 November[33].

Across Northern Europe, at least 16 people were killed[34]. All flights were cancelled from Amsterdam Schiphol airport and there were numerous cancellations from Spanish airports. An estimated 1.2 million households in northern France were left without electricity[35] and more than 1 million residents were cut off from the mobile telephone network. Brest and Quimper Airports were also shut and there was disruption to Eurostar operations[36].

Approximately 10,000 homes in Cornwall were left without power, hundreds of schools were closed and many train services were disrupted by fallen trees. Gusts in Channel Islands ranged from 70-90 knots (36-46 m s$^{-1}$)[35] with a maximum gust of 90 knots (46 ms$^{-1}$) recorded in Alderney at approximately 08 UTC on 2 November[37]. Jersey also experienced a T6 tornado with estimated winds in the region of 161-186 mph (71-83 m s$^{-1}$). Its 8-km track left a trail of destruction and tens of people needed to leave their homes. It is likely that this is the strongest tornado reported in the British Isles since the Gunnersbury tornado in December 1954[38].

The 10-m wind speed and MSLP structure of Ciarán are shown in Fig. 2(a, b) at the times when it impacted the land: 00 and 06 UTC 2 November 2023. State-of-the-art model analyses, such as the IFS analysis used in this figure, represent the best three-dimensional estimates of the actual atmospheric state. The low-pressure centre of the storm tracked along the southern UK coast and the strongest winds (turning cyclonically) occurred in an arc in the southwest quadrant of the storm when the strong winds impacted Brittany and later more directly to the south of the low centre when they impacted the Channel Islands. The 10-m winds weakened significantly over land due to surface friction, no longer reaching the threshold for shading in the figure. They also weakened between the two times shown, with the peak winds falling by about 6 ms$^{-1}$, likely due to a combination of the storm making landfall and having already reached its mature stage. The observed wind speeds, shown by the overplotted colour-filled circles, are consistent with the analysed fields, away from the coastlines but exceed those analysed in some locations, notably some coastal locations and at the narrowest point of the English channel. The winds in these locations will be influenced by local mesoscale processes and so these exceedances are not unexpected given the resolution of the IFS model. The track of the storm, defined as the locations of its minimum MSLP according to IFS analyses, is shown by the black symbols joined by lines in FigurI(c). Ciarán had its genesis in the western North Atlantic around the time of the first track point shown (06 UTC 31 October) and travelled rapidly eastwards across the North Atlantic. The contours shown illustrate the MSLP and 250-hPa wind speed (i.e., the upper-level jet) at the start, middle and end times of the tracks and show how Ciarán evolved from a weak disturbance (with central MSLP exceeding 995 hPa) to a record-breaking deep storm as it crossed from the equatorward to poleward side of the jet at about 06 UTC 1 November.

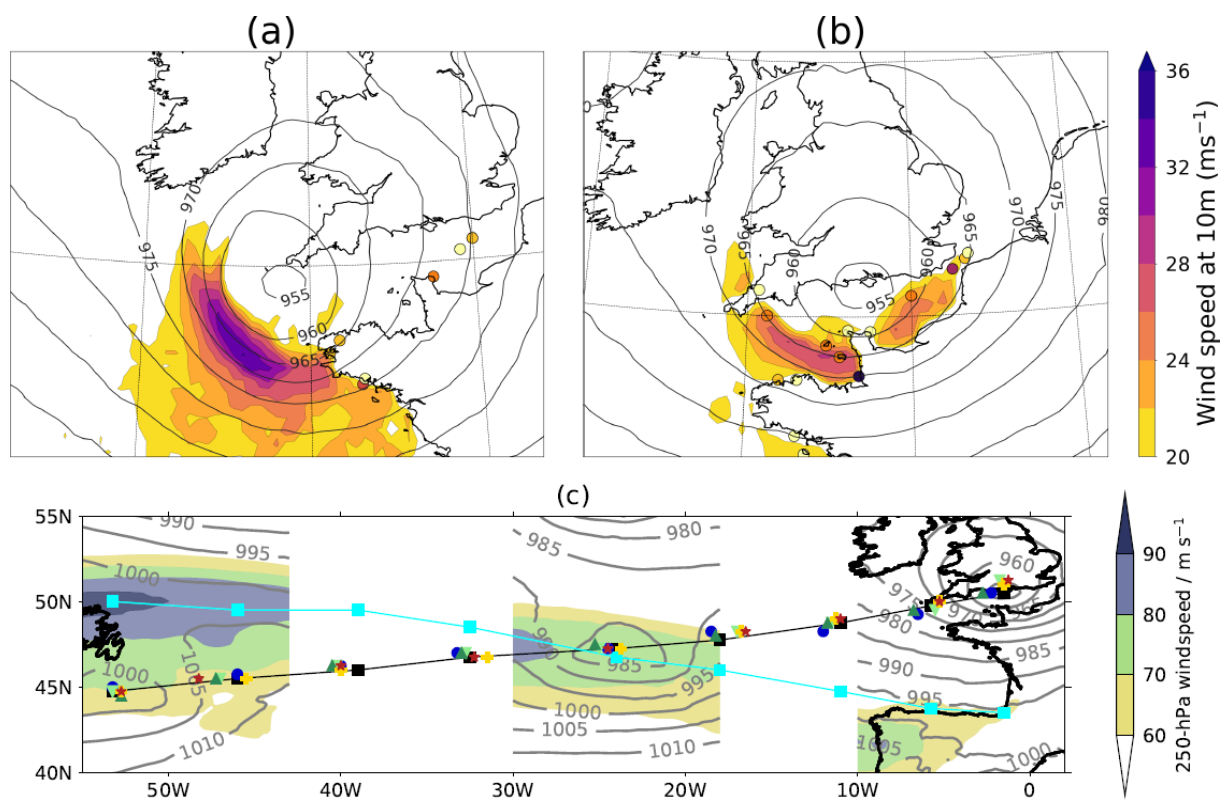

*Figure 22. (a, b) Maps of 10-m wind speed (shading) and MSLP (contours) at (a) 00 UTC and (b) 06 UTC 2 November 2023 from the IFS analysis. Synoptic wind observations above 20 m s$^{-1}$ are shown as coloured dots. (c) Six-hourly track points from the IFS analysis (black squares joined by lines) and the IFS HRES forecasts and AI models (coloured symbols as in Figure 2(a, b)) from 06 UTC 31 October to 06 UTC 2nd November 2023 (left to right) together with partial MSLP (grey contours in hPa) and 250-hPa wind speed (colour-filled contours $^{-1}$) from the IFS analysis at 06 UTC on 31 October, 1 November and 2 November (left to right). The locations of the jet maxima at the longitude points of the MSLP minima at each time are indicated by the cyan squares connected by lines.*

Track, intensification and wind impacts of Storm Ciarán

Ciarán's track was well forecast by both the IFS HRES and ML-models (I. 2(c)) initialised at 00UTC on 31 October, although small differences in the location of the storm centre, and associated wind field, were critical for the accurate predictions of weather warnings along the southern English coast. Two days before Ciarán began to impact land and well before the start of its fast intensification, the spread in the position of the storm in ML-models and NWP-models is similar.

The evolution of the minimum mean sea pressure (MSLP) at the centre of the developing storm and its associated maximum 10-m wind speed are shown in Fig. 3 for the IFS analysis, IFS HRES forecast and ML model forecasts in panels (a, c), and for the ERA5 reanalysis, forecasts based on the ERA5 system, and the control (unperturbed) ensemble members of four NWP models in panels (b, d). Considering first the minimum (MSLP) evolution, all the forecasts closely follow both analysis products, capturing both the rapid deepening phase of the storm and its maximum intensity depth. The minimum MSLP at the end of the forecast (06 UTC 2 November) is 954 hPa in both the IFS analysis and ERA5. This value varies between 951 and 955 hPa for the ML models and between 950 and 953 hPa for the six NWP models (including the IFS HRES). In contrast, the spread in the maximum wind speed evolution is far greater. At the time of peak wind speed in both analyses, 00 UTC 2 November (48-h lead time), the value in the IFS analysis is 34 m s$^{-1}$. The IFS HRES forecast predicts this well (36 m s$^{-1}$), while the other, generally slightly coarser resolution, NWP models mostly forecast slightly weaker winds (30-37 m s$^{-1}$) with the NCEP model being a clear outlier, predicting winds of 40 m s$^{-1}$. The wind speeds forecast by the ML models are far too weak (25-26 m s$^{-1}$), even in comparison with the analysis from ERA5 (30 m s$^{-1}$). The ML models fail to capture the rapid intensification of the winds after about 06 UTC 1 November (30-hour lead time). Forecasts made using the ERA5 analysis system do not suffer from this low wind bias and so the underestimation is unlikely to be the result of training the ML models on the ERA5 data. The economic loss resulting from strong surface winds is often assumed to scale as the cube of normalised wind gust speed over a threshold (such as the 98th percentile value)[39], so even a small underestimation in predicted wind speed can be significant in terms of the subsequent losses.

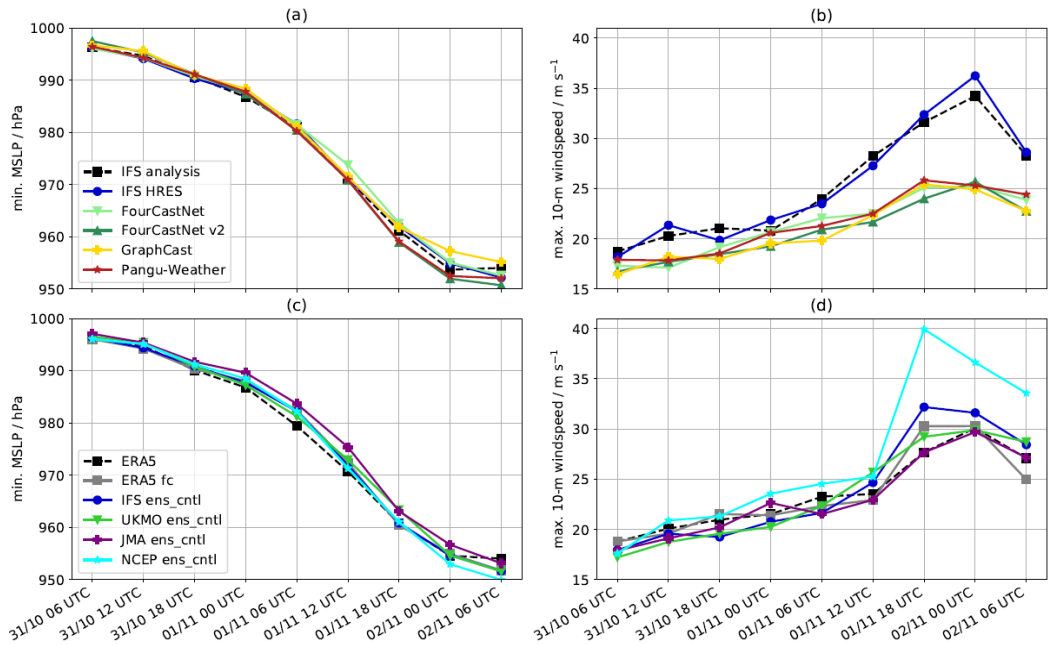

*Figure 33. Timeseries of (a, c) mean sea level pressure and (b, d) 10-m wind speed in analyses and in forecasts starting from 00 UTC 31 October 2023. (a, b) IFS analysis, IFS HRES forecast and forecasts from the four ML models. (a, d) ERA5, forecasts made from the ERA5 system and forecasts from control members of the ensemble forecasts from IFS (IFS ens_cntl), the Met Office (UKMO ens_cntl), the Japan Meteorological Agency (JMA ens_cntl) and National Centres for Environmental Prediction (NCEP ens_cntl). Note that the ECMWF HRES and IFS control member use the same model and resolution but are not bit-identical for technical computational reasons.*

The differences in maximum 10-m wind speed are explored further in Fig. 4 which shows maps of the 10-m wind speed and MSLP for ERA5, the IFS HRES forecast and the four ML-models valid at 00 UTC 2 November, the time of peak wind speed in both analyses and when the strong winds made landfall in France. All the forecasts were initialised 48 hours prior to this time (as for the data shown in Fig. 3). These maps can be compared directly with the IFS analysis fields shown in Fig. 2(a). The region of strong winds is located in an arc in the region of the tight MSLP gradient in the southwest quadrant of the Ciarán in all seven maps. However, the ML models fail to predict the strongest winds in a band following the isobars (contours of constant MSLP) in the region of the tightest MSLP gradient, as is seen the IFS HRES forecast, ERA5 and the IFS analysis. It is notable that, despite all the ML models being trained on ERA5, they fail to capture the structure and magnitude of the winds in ERA5 (including in forecasts made using the ERA5 system, as shown in Fig. S1(a)) for this storm, implying that the far weaker winds found for the ML models compared to the NWP forecasts and IFS analysis are not simply a consequence of them being trained on a coarser resolution dataset. Note the NWP models used in Fig. 2 have a similar resolution to ERA5 (equivalent grid spacings of ERA5 ~31 km, Met Office ~20 km, JMA ~27 km, NCEP ~25 km) with the exception of the IFS (~9 km).

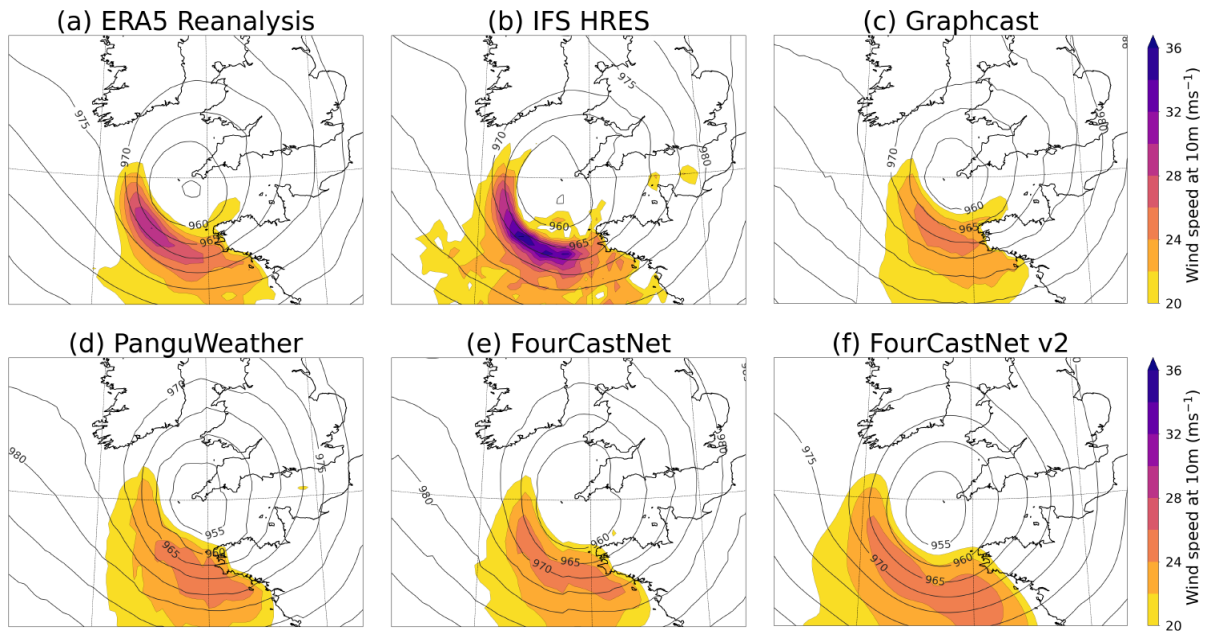

*Figure 44. Maps of 10-m wind speed (shading) and MSLP (contours) at 00 UTC 2 November 2023 from (a) ERA5 and (b-f) forecasts, initialised at 00 UTC 31 October 2023, from the (b) IFS HRES model and (c-f) ML models, as labelled .*

Dynamical structure of Storm Ciarán

In this section, we evaluate the dynamics of Storm Ciarán during the final stage of its rapid development with a focus on the formation of strong winds at low altitudes.  We compare the predictive capability of the ML-models by comparing with the IFS HRES forecast and ERA5. The ML-models are all trained on ERA5 and have the same resolution as the output provided for ERA5, allowing a fair comparison of model performance. The forecasts are all initialised at 00 UTC 1 November, during the onset of Ciarán's rapid intensification phase (Fig. 3(a)).  They are evaluated 18 hours later (Fig. 5) and 24 hours later (Fig. 6), when Storm Ciarán's peak wind speeds were observed. By shifting the focus to these short lead times from the previous section, the aim is to highlight both the similarities in and differences between, the NWP and ML forecasts on timescales relevant for refining hazard warnings. To aid the reader, some key parts of the storm structure are labelled in Fig. 5(a).

On 1 November 2023, Ciarán underwent significant intensification beneath the left exit region of an upper-level jet streak (Fig. 2(c)). All the ML-models captured the position and extent of the upper-level jet streak accurately with the minimum MSLP associated with Ciarán beneath the left exit region at 18 UTC (Fig. 5(a-f)), a critical aspect of Storm Ciarán's dynamics.

There is also consensus among the ML-models concerning the general shape of the cyclone. Figure 5(a)-(f) shows the position of the selected moist isentropes , chosen to indicate the frontal locations and, by their separation, the frontal strengths. The position of the warm sector, identified as the region inside the 285 K moist isentrope, is characterised as a hooked feature in ERA5. The shape of the warm sector is well captured by all the ML-models. The cloud head, represented by 700-hPa relative humidity above 80% (grey shading), is seen wrapping around the poleward side of the cyclone centre in ERA5. The IFS HRES, Graphcast and PanguWeather forecasts accurately depict the shape of the cloud head; however, the cloud head in FourCastNet v2 appears less curved than the other forecasts. FourCastNet forecast does not output a humidity variable at 700 hPa.

Despite capturing the general shape of Storm Ciarán, there are noticeable differences in the strength of frontal structures, indicated by the gradient in wet-bulb potential temperature (how close together moist isentropes are). This is true for the cold front, denoted by the 285-K and 287.5-K moist isentropes to the southeast of the cyclone centre and also for the "bent-back front", i.e., the gradient between the 282.5-K and 285-K moist isentropes that wrap around the cyclone centre on its northwestern side. To the south-west of the low centre the moist isentropes indicating the bent-back front diverge, and this is known as the frontal-fracture region. While all the ML forecasts include a frontal-fracture region, they struggle to resolve the sharp across-front temperature gradient to the west and south-west of the low centre.

Cross-frontal wind shear is another indicator of frontal strength, and the values of the vertical component of 850-hPa relative vorticity (green shading in Fig. 5(a)-(f)) near the bent-back front provide further evidence of the difficulties that the ML-models have in simulating the frontal structures in the region. The hook-shaped narrow strip of high relative vorticity aligned with the bent-back warm front that is present in both IFS HRES and ERA5 (with maximum values up to $9\times10^{-4}$ $s^{-1}$ and $7\times10^{-4}\,s^{-1}$, respectively) becomes broader and weaker in the ML models. The discrepancy between the ERA5 (and also the forecasts based on the ERA5 system, see Fig. S1(b)) and ML models in representing the sharpness of the bent-back front indicates that this shortcoming of the ML models is not solely due to model resolution.

This difference in frontal strength is particularly significant since it directly relates to the environment that can be conducive to the descent of a sting jet. Sting jets are coherent air flows that descend over few hours from inside the tip of the cloud head at mid-tropospheric levels leading to a distinct mesoscale (perhaps 50-100 km across) region of near-surface stronger winds, and particularly gusts[40]. Among the models presented in this study, only the IFS forecasts and analysis have the necessary resolution to resolve a sting jet. It is crucial to recognise that the ML-models, trained on coarse resolution data, are not equipped to discern features such as the presence of mesoscale jets. This limitation highlights the importance of using models with adequate resolution when predicting high-impact weather phenomena occurring at small spatial scales. However, our analysis suggests that the ML-models struggle to represent frontal structures conducive to mesoscale high-impact features even when compared against NWP models with similar resolution, such as that used to generate ERA5.

The lack of sharpness of the bent-back warm front and cold front in the ML models impacts the strength of the wind speed maxima, as can be seen by turning the focus of evaluation to the region of strong winds. Figure 5(a)-(f) shows the 850-hPa wind speed (filled contours) to give an indication of the lower-tropospheric storm structure that is less influenced by the presence of land below than the 10-m winds shown previously; consequently, winds at this pressure level, roughly a km above the ground, are normally stronger than those nearer the surface. All the ML models consistently identify two regions of strong winds: one in the frontal-fracture region and another in the warm sector. The strong winds situated in the frontal-fracture region are associated with the tight pressure gradient near the tip of the bent-back warm front, the associated descent and acceleration where the moist isentropes spread out, and the alignment with the direction of propagation of the storm (with a possible local enhancement due to sting-jet descent in the IFS HRES, see the small-scale areas above 46 ms$^{-1}$). The strong winds mostly in the core of the warm sector (enclosed by the 287.5-K moist isentropes) are associated with a broad jet, known as the warm conveyor belt jet, which ascends through the depth of the atmosphere from the top of the atmospheric boundary layer. The ability of the ML models to identify both the frontal-fracture and warm conveyor belt wind maxima (albeit with differences in the spatial structure and intensity of the latter) underscores their ability to accurately capture the general structure of extratropical cyclones. However, maximum wind speeds are weaker in the ML models than in ERA5, where they exceed 46 m s$^{-1}$ (and even 48 m s$^{-1}$ in the forecast based on the ERA5 system) in a broad region at the entrance of the frontal fracture. While Graphcast and FourCastNet display a small deficit of around 2 m s$^{-1}$, PanguWeather and FourCastNetv2 are roughly 4 m s$^{-1}$ and 6 m s$^{-1}$ lower, respectively.

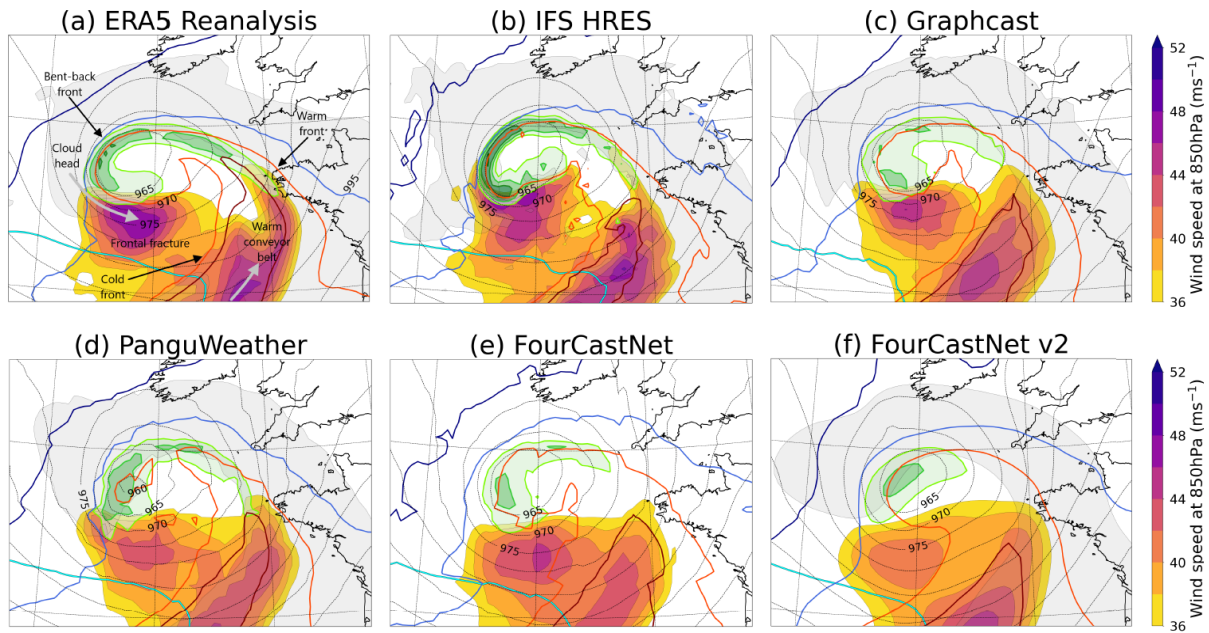

*Figure 5: Maps of wind speed at 850 hPa (shading), wind speed at 250 hPa (65 ms$^{-1}$, cyan contour with high values in the bottom left of the panels), wet-bulb potential temperature at 850 hPa ( dark blue, light blue, light red and dark red contours indicating values increasing every 2.5K from 280K to 287.5K), MSLP (thin grey contours), relative humidity with respect to water at 700 hPa (grey shading encircling regions above 80%, not shown for FourCastNet (e) as not available), vertical component of relative vorticity at 850 hPa (light-to-dark green shading , from 3 x 10$^{-4}$ s$^{-1}$ and then every 2 x 10$^{-4}$ s$^{-1}$). All maps refer to 18 UTC 1 November 2023. The IFS high resolution forecast (b) and forecasts from the four ML models (c-f) all started at 00 UTC 1 November 2023, while ERA5 (a) is a reanalysis dataset. Note that the range of the wind speed colour bars in Figs. 5 and 6 is different to that in Figs. 2 and 4. The main features of the cyclone described in the text are annotated in panel (a).*

We now turn our attention to the forecasts valid 6 hours later, at 00UTC on 2 November 2023 (Fig. 6(a)-(f)), the time of peak ERA5 wind speeds. At this stage of Ciarán's evolution, analysis of the frontal-fracture region and the warm sector reveals several interesting features. ERA5 exhibits a region of warm air at the centre of the storm, which is separated from the main warm sector, a process known as warm core seclusion. Warm core seclusion occurs during the mature stage of extratropical cyclone development when cold air wraps around the low centre and cuts it off from the warm subtropical airmass. Relative vorticity starts decreasing along the bent-back front and generally increases near the cyclone centre as the front wraps around it. While the general evolution is captured by all models, the degree of clarity in the presence of a well-defined warm seclusion varies noticeably among the ML-models.

Focusing on the maximum winds in the frontal-fracture region, now compounded by the arrival of the cold conveyor belt (the main low-level jet in the cold sector, behind the cold front, of an extratropical cyclone), reveals that the ERA5, the forecast based on the ERA5 system (Fig. S1(c)) and the IFS forecast all have peak 850-hPa wind speeds between 48-50 m s$^{-1}$. GraphCast and FourCastNet exhibit peak wind speeds 4-6 m s$^{-1}$ lower. PanguWeather and FourCastNet-v2 have a larger weak bias, with wind speeds underestimated by 6-8 m s$^{-1}$. Wind maxima are consistently underestimated in the ML models when compared to the benchmarks provided by the ERA5 (and its forecast) and the IFS forecast. This discrepancy in predicting wind maxima at the time of peak winds and as they approach land is a crucial for assessing the potential impact of Storm Ciaran's surface winds and associated gusts.

By inspection, the structures of the MSLP fields in Figs. 5 and 6 are similar for the different models despite the differences in the wind speed structure and magnitude. This raises a further interesting question, is the discrepancy between the wind maxima in the conventional NWP and ML because the ML models do not reproduce the dynamical balances between the wind and pressure fields inherent in the conventional NWP models? This question is examined in detail in the Supplementary Material (and included Figs. S2-8) and a short summary is included here.

While the calculation of geostrophic wind field (resulting from the balance the pressure gradient and Coriolis forces) is relatively straightforward, calculating the more accurate gradient wind field (with the further inclusion of the centrifugal force associated with the curvature of a parcel trajectory) is more complex. Since both calculations require the evaluation of horizontal gradients in the geopotential field, an unphysical lack of smoothness on the smallest scales in all of the ML models becomes easily apparent and should be further investigated. Note that while the gradient wind should provide a better approximation to the frictionless large-scale flow (where that flow is curved) than the geostrophic wind, the flow can still differ from gradient wind balance due to unbalanced motions which may be physically realistic, particularly in high resolution model output (and reality).

Smoothed geostrophic and gradient wind fields have physically plausible structures in both NWP and ML model outputs. While the strongest full winds are found in the NWP model outputs even after smoothing, the strongest gradient winds are not clearly different between the ML and NWP models. The differences between the smoothed full and gradient wind fields for the NWP and ML models have similar characteristic structures and magnitudes in strong wind regions of the storm. Within the limitations of the accuracy of our calculations, we cannot conclude that the weak winds in the ML model forecasts are the result of an inability to resolve the proper dynamical balances, but are likely to instead be related to inadequacies in the geopotential field, i.e., in the gradient and curvature of the geopotential contours.

In summary, the ML-models represent the large-scale dynamical drivers key to the development of Storm Ciarán well, including the position of the storm relative to the upper-level jet exit. They also accurately capture the larger synoptic-scale structure of the cyclone such as the position of the cloud head, shape of the warm sector and location of the warm conveyor belt jet. The ability of the ML models to resolve the more detailed structure of the storm is more mixed. Only some ML models correctly resolve the warm core seclusion and none of them capture the sharp bent-back warm frontal gradient. ML models underestimate the magnitude of the strongest winds at the surface and in the free atmosphere (above the boundary layer), particularly in the frontal-fracture region near the end of the bent-back front. Note that this underestimation of the strongest wind speeds is not a consequence of the resolution of the output of the ML models or their training data, since it also applies when comparing against the ERA5 (and forecasts based on the ERA5 system) and NWP models with resolution similar to ERA5.

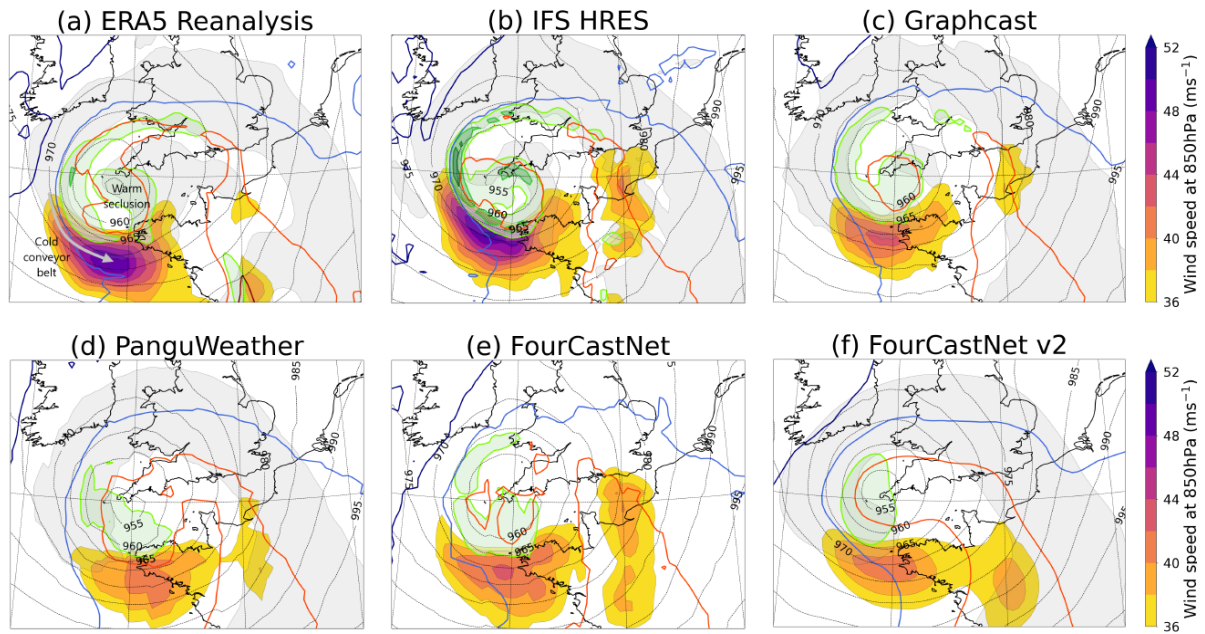

*Figure 6: As Figure 4 but for 00 UTC 2 November 2023. ⁺ Contours of wind speed at 250hPa and some of the contours of wet-bulb potential temperature at 850 hPa are not present as the associated values are not reached in the maps shown.*

**Discussion**

The contrasting ability of the four ML models considered to accurately forecast the large-scale dynamical properties of Storm Ciarán and its damaging winds serve to highlight the need for more comprehensive assessment of this new and potentially transformative forecasting tool. More than 48 hours before Storm Ciarán affected communities surrounding the English Channel, forecasts of the rapid MSLP deepening and track of the storms produced by the ML-models were essentially indistinguishable from forecasts from an ensemble of conventional NWP models. Our analysis shows that the ML models were able reproduce the upper-level flow that steered the developing storm into the left exit region of the jet and led to its rapid intensification. Many of the important dynamical features of the storm including the position of the cloud head, shape of the warm sector and location of cold and warm conveyor belt jets were also well captured by the ML models. The ML models do not seem to have been limited by the fact that a storm of comparable central pressure has never previously been observed over England during November. However, even in the relatively short ERA5 record, storms developing in a similar way with similar dynamical drivers (such as the upper-level jet) are common throughout winter and the ability of ML models to forecast more dynamically unusual storms, such as small-scale storms that develop rapidly from waves on pre-existing fronts, is an open question.

In contrast, when considering the damaging winds associated with Storm Ciarán in detail, forecasts from the ML models had significant errors and poorer performance than conventional NWP models. All four ML models failed to produce the narrow band of very strong winds at the surface that led to the most severe impacts, The ML-models also failed to represent the strength of the cross-front thermal gradient in the bent-back front (a feature also dynamically linked to strong winds) and had variable success in producing the warm seclusion of air that formed in the centre of the storm in its mature stage.

Much further work, considering other storms, is needed to assess if the biases apparent in the simulation of Storm Ciarán are a systematic feature of this first generation of ML models. Increased scrutiny of the models is likely to lead to the identification of target areas for model improvement, as it has done for NWP models. Since the ML models are available to all through public repositories, this scrutiny is likely to enable rapid model improvement. Detailed documentation of the performance of ML models will also be critical to weather forecasters seeking to make greater use of the ML models as part of the forecasting process. Forecasting centres like ECMWF are already beginning to develop and test alpha versions of ML models that complement their existing capabilities[41].

Based on a single case study, it would be premature to draw conclusions about the relative abilities of the four different approaches to ML weather forecasting exemplified by the different models. In particular, given we only had access to the 'small' version of the FourCastNet-v2 model it might be expected that this model would have a limited ability to produce the detailed properties of Storm Ciarán. Nonetheless, studies like ours are useful to identify knowledge gaps in ML model development for forecasting, particularly in their ability to capture the structure of extreme weather patterns. This can be direct, such as the inclusion of necessary output variables (a 700-hPa humidity variable to identify the shape of the cloud head missing from the FourCastNet model), or through the formulation of more nuanced hypotheses for investigation. For instance, PanguWeather's ability to capture the vertical component of the 850-hPa relative vorticity could be due to the model integrating height information across levels. Similarly, GraphCast's ability to simulate the warm core seclusion better may be due to its multi-mesh representation rather than the spatial mixing used in the other models. All the ML models failed to capture the intense winds – both at the surface and in the lower troposphere – associated with Ciarán, and this may indicate that future models could benefit from including variables and the relationships across these variables that better characterize the planetary boundary layer in their training datasets. Most importantly, our analysis makes a strong case for a more robust evaluation of the forecasts from the ML-models across all relevant spatio-temporal features of the physical phenomenon considered instead of isolated error metrics on individual variables.

The rapid acceleration of the forecasting capabilities of ML-models as exemplified by our study of Storm Ciarán poses many new challenges and opportunities for atmospheric science[42]. Explainable AI (xAI) techniques[43,44] could be powerfully combined with the ML-models we have considered to develop a deeper understanding of the reasons that they were able to produce skillful forecasts of Storm Ciarán in line with other attempts to unify ML and causal discovery methods[45]. The development of general-purpose, foundational models[46] could add further to the set of tools available to both forecasting and understand high-impact weather events.

**Methodology**

In this study we compare forecasts produced by four different models based on machine learning methods. All are initialised from the same operational ECMWF analysis allowing a direct comparison with the current operational forecast of the ECMWF high-resolution model (CY48R1). The ML-model forecasts are produced using the ai-models toolbox developed by ECMWF (https://github.com/ecmwf-lab/ai-models).

All four models considered are data driven Deep Learning models and originally trained on a few (~4) decades worth of atmospheric and surface variables from the ERA5 dataset[47] at a resolution of 0.25°x 0.25° (~30 km resolution at the equator), which translates to 720 x 1440 grid cells. The ML models are all autoregressive, which means model output from a given time step can be used to predict output at the next time step. Model differences arise chiefly from the individual architectures, the selection of variables, parameterizations and training schemes briefly summarized in the technical details below:

- FourCastNet[12], uses the vision transformer (ViT) architecture with an Adaptive Fourier Neural Operator (AFNO)[48]. The AFNO enables dependencies across spatial and channel dimensions to be modelled efficiently at high resolutions where spatial token (feature) mixing occurs as a global convolution in the Fourier domain with FFTs. The model has a pre-training step where the AFNO is trained ahead on the ERA5 data with 20 different surface and atmospheric variables and then used for inference. The pre-training step learns mappings between **X**(t) and **X**(t+ Δt) where t is a time step, Δt is a time increment (set to 6h) and **X** is a tensor of features called patches. In the second fine-tuning or inference step, the pre-trained model is used to produce inferences from a defined state **X**(t), first for **X**(t + Δt ) and this output from the models is itself then used to generate **X**(t + 2* Δt) or the output for the second time step. Thus, while the training of the model is resource intensive, it is a one-time cost and the inference step is very fast.
- FourCastNet v2[13] is a development of the original FourCastNet model that uses Spherical Harmonics Neural Operators for modeling non-linear chaotic and dynamical systems on a sphere as opposed to flat Euclidean spaces. The model is trained with ERA5 data in a two-step process similar to FourCastNet – a single autoregressive step followed by fine tuning. By learning global convolutions in computationally efficient manners, Fourier Neural Operators (such as those used in FourCastNet) are capable of accurately simulating long-range dependencies in spatio-temporal data. However the Discrete Fourier Transform that FNOs rely on assume a flat geometry, resulting in dissipation together with visual and spectral artefacts. The Spectral Fourier Neural Operators (SFNO), forming the basis for the FourCastNet v2 model architecture in its update from the FourCastNet model, in addition to having the desirable properties of FNOs also have translational or rotational equivariance. FourCastNet v2 is trained on a 73-channel subset of the ERA5 reanalysis dataset on single levels and pressure levels.

- Pangu-Weather[14] consists of four deep neural networks with different lead times (time between input and output) of 1h, 3h, 6h and 24h. 5 upper atmosphere and 4 surface variables at 13 different pressure levels were used to train the model with a combined total of 256 million parameters. The overall deep network architecture is called 3DEST or 3D-Earth specific Transformer that integrates height information into a new dimension thus capturing relationships between atmospheric variables across pressure levels unlike similar transformer-based models such as FourCastNet. Data is fed into the neural network and a process called patch embedding is used to downsample the input data from individual grid cells into a 3D cube. This cube is then put through an encoder-decoder based on a ViT called the Swin transformer[25] with 16 blocks. The positional bias in the Swin transformer is replaced with an Earth-specific positional bias to reflect the fact that in a 2D projection of a sphere, distances between neighboring points is not the same across all latitudes. The decoder is symmetric to the encoder. Although the transformer based neural network has a large training time similar to FourCastNet, this is partially improved in Pangu-Weather by the use of an hierarchical temporal aggregation scheme that reduces cumulative forecast errors and also the forecast generation time. This is done by employing the neural network with the largest lead time iteratively for a forecast so that neural networks with shorter lead times are used closer to the forecast. The height integration and aggregated forecast schemes are also considered improvements over other transformer-based architectures.

- GraphCast[15] is based on Graph Neural Networks (GNNs)[49] with around 36.7 million parameters. The model is trained with 5 surface and 6 atmospheric variables at 37 pressure levels resulting in 227 variables for every data point or grid cell. In the first step, the Encoder maps information from individual grid cells to nodes in a multi-mesh representation. The multi-mesh is derived as icosahedral meshes of increasing resolution from coarse (12 nodes) to fine (40,962 nodes). The second step has Processors using 16 GNN layers to propagate local and long-range information across the nodes on the multi-mesh through message passing. Finally, the decoder uses a single GNN layer to map the final processor layer's multi-mesh representation back to the grid cells. GraphCast thus avoids the use of transformers and the associated scaling issues with higher resolutions that could result in large training times.

*Numerical weather prediction model forecasts and analysis products.*

The ML model forecasts are compared to a set of forecasts from conventional numerical weather prediction (NWP) models to assess both systematic differences in the capabilities of the NWP and ML models and how the spread in the forecasts from the two architectures compare. Forecasts from the IFS HRES forecast and forecasts based on the ERA5 system (see below for description of ERA5) were obtained from ECMWF and control (unperturbed) members of the ensemble forecasts for four models (the IFS[50], the Met Office[51], the Japan Meteorological Agency (JMA)[52], and the National Centres for Environmental Prediction (NCEP)[53]) were downloaded from the TIGGE archive[54] of operational global ensemble weather forecasts out to medium range. The models chosen differ in their design and the resolution of the numerical model grid. Cycle 48r1 of the IFS was operational at the time of Storm Ciarán. Following the upgrade to this cycle in June 2023, the HRES and ensemble forecasts have the same resolution, equivalent to 9 km grid spacing. The Met Office, JMA and NCEP ensembles have grid spacings of approximately 20 km, 27 km and 25 km, respectively. The data for the four control ensemble members were all obtained regridded to a regular latitude-longitude grid of 0.5 degrees.

*Analysis products*

Both sets of models are compared to two analysis products (optimal blends of short-range forecasts and observations): the operational IFS analysis and ERA5[47]. The operational IFS analysis is produced using the IFS HRES forecast and has a resolution equivalent to 9 km grid spacing, ERA5 has a resolution equivalent to 31 km grid spacing. The IFS analysis and ERA5 were regridded to a regular latitude-longitude grid of 0.25 degrees.

ERA5 is used as an additional measures of forecasts 'truth' because the ML-models all used ERA5 as their training data. Hence comparison with ERA5 indicates the skilfulness of these models relative to the best possible forecast given their training data. It is to be expected that the IFS analysis will include smaller-scale and higher amplitude weather features than ERA5 due to the use of a higher resolution model, despite being regridded to the same grid. It is also expected that the IFS analysis will be closer to the "truth" due to the use of higher resolution and an upgraded modelling system.

**Acknowledgments**

We thank the ECMWF labs team for building the publicly available ai-models library which enabled us to produce and compare forecasts from the four ML models. This library can be accessed at https://github.com/ecmwf-lab/ai-models. We also thank the modelling groups who made the code for the ML models publicly available through the following repositories:

- FourCastNet https://github.com/NVlabs/FourCastNet
- PanguWeather https://github.com/198808xc/Pangu-Weather
- GraphCast https://github.com/google-deepmind/graphcast

This work is partly based on TIGGE data. TIGGE (The International Grand Global Ensemble) is an initiative of the World Weather Research Programme (WWRP). We are also grateful to ECMWF for providing access to operational analysis products to members of the research team through national research accounts held through the Met Office.

The work is partly funded by the UKRI Natural Environment Research Council (UKRI-NERC) through several grants held by contributors and by the Schmidt Futures Foundation.

BH is funded by the UKRI-NERC CANARI programme (NE/W004984/1).

NJH is funded by UKRI-NERC UMBRELLA (NE/X018555/1).

KMRH is funded by a UKRI-NERC Independent Research Fellowship (MITRE; NE/W007924/1).

RS is funded by UKRI-NERC TerraFIRMA (NE/W004895/1).

AV is funded by UKRI-NERC Arctic Summer-time cyclones (NE/T006773/1).

SD is funded and supported by the Schmidt Futures Foundation.

**Author Contributions**

AJCP conceived the analysis and produced ML forecasts and initial characterisation of their performance. SD, KMRH, RS and RV contributed to the analysis and description of the ML models. NJH led analysis of the properties of Storm Ciarán and its impacts. HD, SG, BH and AV produced the detailed dynamical analysis of the properties of Storm Ciarán in the forecasts including collecting NWP data and producing all figures.

All authors contributed to discussions, writing, proof reading and editing the manuscript.

All authors should be considered co-first author as the work was completed collaboratively.

The funders played no role in study design, data collection, analysis and interpretation of data, or the writing of this manuscript.

**Data Availability**

The datasets used and/or analysed during the current study available from the corresponding author on reasonable request.

**Code Availability**

Apart from the python packages referenced in the Acknowledgements, the underlying code for this study is not publicly available but may be made available to qualified researchers on reasonable request from the corresponding author.

**Competing Interests**

The Authors declare no Competing Financial or Non-Financial Interests.

**Figure legends**

Figure 5. Surface land and ship station SYNOP observations of Storm Ciarán at 06 UTC 2 November 2023 extracted from the MetDB database[27,28], which holds data including surface and upper air observations and some satellite data. The observations are shown as simplified station circles using conventional notation[29]. Circle shading indicates cloud cover in octas, wind barbs and feathers indicate wind speed in knots with the wind direction towards the circle, and numbers are the last three digits, including a decimal place, of the MSLP (in hPa) e.g., 543 equates to 954.3 hPa. Some thinning of observations has been performed for clarity and note that two ships both reported at 51.1°N, 1.7°E with different wind speeds and directions.

Figure 26. (a, b) Maps of 10-m wind speed (shading) and MSLP (contours) at (a) 00 UTC and (b) 06 UTC 2 November 2023 from the IFS analysis. Synoptic wind observations above 20 m s$^{-1}$ are shown as coloured dots. (c) Six-hourly track points from the IFS analysis (black squares joined by lines) and the IFS HRES forecasts and AI models (coloured symbols as in Figure 2(a, b)) from 06 UTC 31 October to 06 UTC 2nd November 2023 (left to right) together with partial MSLP (grey contours in hPa) and 250-hPa wind speed (colour-filled contours) from the IFS analysis at 06 UTC on 31 October, 1 November and 2 November (left to right). The locations of the jet maxima at the longitude points of the MSLP minima at each time are indicated by the cyan squares connected by lines.

Figure 37. Timeseries of (a, c) mean sea level pressure and (b, d) 10-m wind speed in analyses and in forecasts starting from 00 UTC 31 October 2023. (a, b) IFS analysis, IFS HRES forecast and forecasts from the four ML models. (a, d) ERA5, forecasts made from the ERA5 system and forecasts from control members of the ensemble forecasts from IFS (IFS ens_cntl), the Met Office (UKMO ens_cntl), the Japan Meteorological Agency (JMA ens_cntl) and National Centres for Environmental Prediction (NCEP ens_cntl). Note that the ECMWF HRES and IFS control member use the same model and resolution but are not bit-identical for technical computational reasons.

Figure 48. Maps of 10-m wind speed (shading) and MSLP (contours) at 00 UTC 2 November 2023 from (a) ERA5 and (b-f) forecasts, initialised at 00 UTC 31 October 2023, from the (b) IFS HRES model and (c-f) ML models, as labelled.

Figure 5: Maps of wind speed at 850 hPa (shading), wind speed at 250 hPa (65 ms$^{-1}$, cyan contour with high values in the bottom left of the panels), wet-bulb potential temperature at 850 hPa (dark blue, light blue, light red and dark red contours indicating values increasing every 2.5K from 280K to 287.5K), MSLP (thin grey contours), relative humidity with respect to water at 700 hPa (grey shading encircling regions above 80%, not shown for FourCastNet (e) as not available), vertical component of relative vorticity at 850 hPa (light-to-dark green shading, from 3 x 10$^{-4}$ s$^{-1}$ and then every 2 x 10$^{-4}$ s$^{-1}$). All maps refer to 18 UTC 1 November 2023. The IFS high resolution forecast (b) and forecasts from the four ML models (c-f) all started at 00 UTC 1 November 2023, while ERA5 (a) is a reanalysis dataset. Note that the range of the wind speed colour bars in Figs. 5 and 6 is different to that in Figs. 2 and 4. The main features of the cyclone described in the text are annotated in panel (a).

Figure 6: As Figure 4 but for 00 UTC 2 November 2023. Contours of wind speed at 250hPa and some of the contours of wet-bulb potential temperature at 850 hPa are not present as the associated values are not reached in the maps shown.

**Supplementary material for Charlton-Perez et al. (2024)**

**Forecasts from the ERA5 dataset**

For completeness we add an additional Figure S1 which shows the structures of Storm Ciarán from forecasts based on the ERA5 system. These are shown to highlight the performance of a conventional NWP system with the same resolution as the training data used for the ML models. ERA5 forecasts are included in the quantitative comparison of Fig. 3.

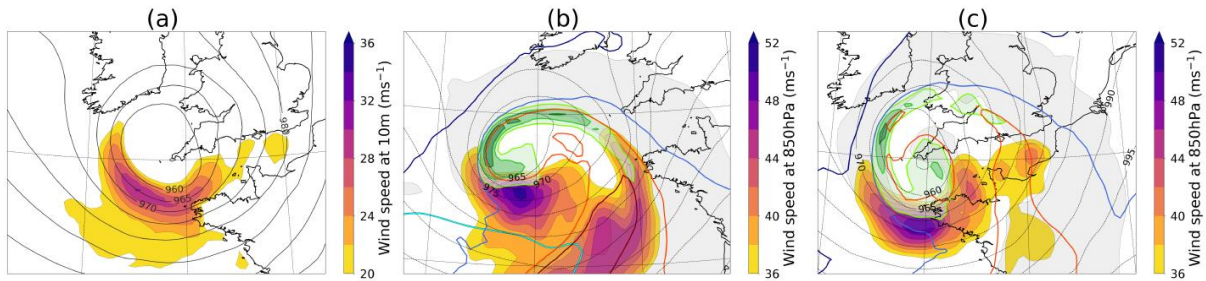

*Figure S1: As in (a) Fig.4, (b) Fig.5, (c) Fig.6 but for forecasts based on the ERA5 system, initialised at (a) 00 UTC 31 October 2023 and (b,c) 00 UTC 1 November 2023.*

**Did ML-models capture the dynamical balance between the wind and geopotential height fields?**

In this section, we outline the calculations performed to establish the extent to which the wind and geopotential fields from the ERA5 Reanalysis, the IFS HRES and the ML models remain close to a dynamically balanced state during the final stage of Storm Ciarán's rapid development. This discussion was prompted by an interesting exchange with two of the reviewers of the submitted manuscript. The key results, stated in the main manuscript, are:

1. The geopotential fields from several ML models contain an unphysical lack of smoothness resulting in noisy geostrophic and gradient wind fields.
2. If the geopotential fields are first smoothed, the resulting geostrophic and gradient wind fields have physically plausible structures in all model outputs. While the strongest full winds are clearly found in the NWP models (ERA5 and IFS HRES), even after smoothing, the strongest gradient winds are not clearly different between the ML and NWP models.
3. Furthermore, the difference between the full and gradient wind forms a similar pattern of sub- and super-gradient winds in the vicinity of the strong full winds region in each model, meaning we cannot conclude the dynamical balance exhibited by the ML models is different to than that of the NWP model, at least given the methodology used here and the complexities of the flow in this case.

To reach these conclusions, we use instantaneous geopotential on the 850 hPa pressure surface to compute the 850 hPa geostrophic wind speed and an estimate of the 850 hPa gradient wind speed for each model, as described below. The choice of the 850 hPa level is made because it is near the typical top of the boundary layer (in the absence of high topography) and so the effects of frictional processes should be small; the winds, wet bulb potential temperature and vertical component of relative vorticity at this level are shown in Figs. 5 and 6 in the main article.

<u>Evaluation of the geostrophic wind</u>

The geostrophic wind $(u_g, v_g)$ is defined as the horizontal wind having the same direction as contours of constant geopotential on a pressure surface but with a magnitude consistent with a balance between the pressure gradient force and the Coriolis force:

$$f \begin{pmatrix} u_g \\ v_g \end{pmatrix} = \begin{pmatrix} -\Phi_y \\ \Phi_x \end{pmatrix},$$

(1)

where $f = 2\Omega \sin \phi$ is the Coriolis parameter, $\phi$ is latitude and $\Phi$ is the geopotential. The derivatives of the geopotential field are evaluated using finite differences.

Figures S2 to S7 show wind fields for ERA5, and forecasts from the IFS HRES and four ML models valid at 18 UTC 1 November 2023 (each figure is for a different model); this time, also used for Fig. 5 in the main article, was chosen as being prior to strong interaction of Storm Ciarán with land although the general conclusions discussed here also apply at 00 UTC 2 November 2023 (not shown), the time used for Fig. 6 in the main article. Panels a and b show the full and geostrophic wind speed at 850 hPa, respectively. The geopotential fields from several ML models contain an unphysical lack of smoothness resulting in noisy geostrophic wind fields. To aid comparison between the models, we smooth the wind components and geopotential using a T106 truncation (≈1 degree resolution in physical space). After smoothing, the full wind speed maps (panels e) contain less small-scale structure and the unphysical noise in the geostrophic wind speed maps (panels f) is removed. From now on we restrict discussion to these smoothed fields.

To compare between models, Figure S8 (a and b) summarises the peak (smoothed) wind speeds and the peak (smoothed) geostrophic wind speeds, defined as the maximum in each of the cold conveyor belt (CCB) and warm conveyor belt (WCB) regions (see annotations in Fig. 5 and 6 for the locations of these features, and Fig. S2-S7 for the precise regions used to evaluate the peak values in S8). The peak full wind is stronger in both regions in the NWP models than the ML models, confirming this finding of the main paper also applies to the smoothed wind field. The strongest peak geostrophic winds occur in the NWP models in both regions, with IFS HRES strongest in the CCB and ERA5 strongest in the WCB, but there is a less clear distinction between the ML models in both cases than for the full wind.

All the models have strong geostrophic winds in the cold sector, far exceeding those in the warm sector. This is despite the smoothed warm conveyor belt winds being similar too, or even stronger, than those in the cold sector and is a consequence of the strong curvature of the flow in the cold conveyor belt region, which is not accounted for in Eq. (1). To better estimate the balanced wind in this region, we next compute an estimate of the gradient wind.

Evaluation of the gradient wind

The gradient wind[1], $(u_{gr}, v_{gr})$, is defined as the horizontal wind having the same direction as the geostrophic wind but with a magnitude consistent with a balance of three forces: the pressure gradient force, the Coriolis force, and the centrifugal force associated with the curvature of a parcel trajectory. As such, it represents a more accurate balance than the geostrophic wind (1). Its magnitude $V_{gr} = \sqrt{u_{gr}^2 + v_{gr}^2}$ satisfies

$$KV_{gr}^2 + fV_{gr} - fV_g = 0,$$

(2)

where $K$ is the curvature of a parcel trajectory. We estimate $V_{gr}$ as outlined below and consider the difference $V - V_{gr}$ as a measure of the extent to which the geopotential field is consistent with the full wind field (i.e., the extent to which the full wind field is in balance).

There is ambiguity in the literature over the choice of parcel trajectory used to compute $K$[1]. Different results arise if the full or the gradient wind (i.e., geopotential contours) are used and if steady or non-steady assumptions are made. Using the full wind to estimate $K$ is dynamically appealing since it provides the most accurate estimate of the true flow curvature. However, in the present case we aim to evaluate the relationship between the wind and geopotential fields in the ML models. Using the wind field to evaluate the curvature which then influences the gradient wind is a circular argument and we opt instead for the simpler geopotential contour approach in which the gradient wind is estimated solely from the geopotential field. This approach is also by far the most commonly used approach in the literature.

To this end, we follow the 'steady contour' method[1] with a modification to include a correction for cyclone movement. To compute the curvature of instantaneous geopotential contours, we rearrange the definition of geostrophic vorticity in natural coordinates to give:

$$K = \frac{1}{V_g}\left(\zeta_g + \frac{\partial V_g}{\partial n}\right),$$

(3)

Where $\zeta_g = v_{g,x} - u_{g,y}$ is the geostrophic relative vorticity and $n$ is distance perpendicular to geopotential contours. We evaluate $K$ from the T106 geopotential field in all cases due to the noise induced by taking multiple derivatives on the raw 0.25-degree data. Due to the fast propagation of the system, a better estimate of the geopotential curvature could be obtained by including information about the time derivative of the geopotential field (the 'non-steady contour' method[1]).

However, the six-hourly frequency of the data available from the ML models is not sufficient for this purpose. Instead, we estimate the instantaneous geopotential tendency by assuming a constant propagation velocity of the cyclone[2] which we estimate from the ERA5 charts at 18 UTC on 1 November and 00 UTC on 2 November as 17.7 m/s in the zonal direction and 6.4 m/s in the meridional direction. The results presented here are not sensitive to minor changes in the choice of this velocity.

Figures S2 to S7 show the gradient wind (panels c) and the full – gradient wind differences (panels d) computed from the non-smoothed geopotential, and the same fields computed from the smoothed geopotential (panels g and h). The gradient winds are much closer to the full winds than the geostrophic winds are. However, as with the geostrophic wind, the gradient wind contains substantial unphysical noise in several of the ML models, so we focus discussion on the smoothed versions from now on, although many of the features are clearly visible in both. The difference between the full and gradient winds forms a characteristic pattern in all the models in the vicinity of the strong full winds. This pattern is a horseshoe-shaped structure of super-gradient winds, with peaks in the region of the warm conveyor belt jet and on the northwards edge of the strong cold sector winds, surrounding a sub-gradient core.

In summary, the IFS HRES is closer to ERA5 than the ML models in terms of the geostrophic and gradient winds (even after smoothing), implying that this NWP forecast best represents the curvature and gradients of the geopotential field, and consequently the associated winds.

From the analysis presented here, we cannot conclude that the ML models exhibit a worse dynamical balance between the wind and geopotential fields than the NWP ones. However, we are limited by the complexity of the flow in this case and the difficulties in calculating balance from the limited data availability. Further work should explore this question in more detail using more accurate balance methods and a variety of storm cases.

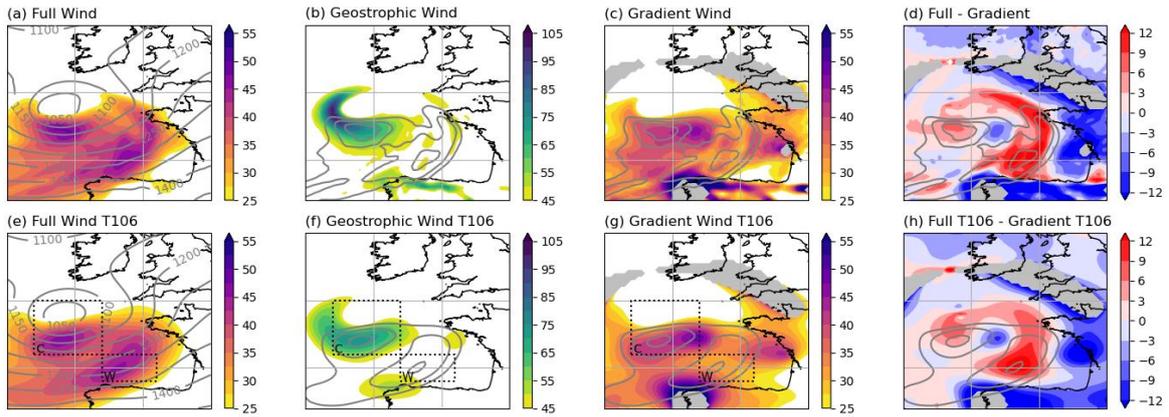

*Figure S2: Maps of 850 hPa wind speed (a), geostrophic wind speed (b), gradient wind speed (c, as described in the text), and the full wind speed minus gradient wind speed difference (d) at 18 UTC 1 November 2023 from ERA5. The bottom row (e-h) shows the same fields computed from the smoothed T106 wind component and geopotential fields. Grey contours are geopotential height (a and e; units: m) and full wind speed (b-d and f-h; contours at 35, 40 and 45 ms$^{-1}$). Dashed boxes indicate the regions used to locate the cold conveyor belt (labelled 'C') and warm conveyor belt (labelled 'W') peak winds used in Figure S7, and the grey shading in (c-d and e-f) mask regions where the gradient wind solution is not defined. Note that the range of the wind speed colour bars is different to those used in the main text.*

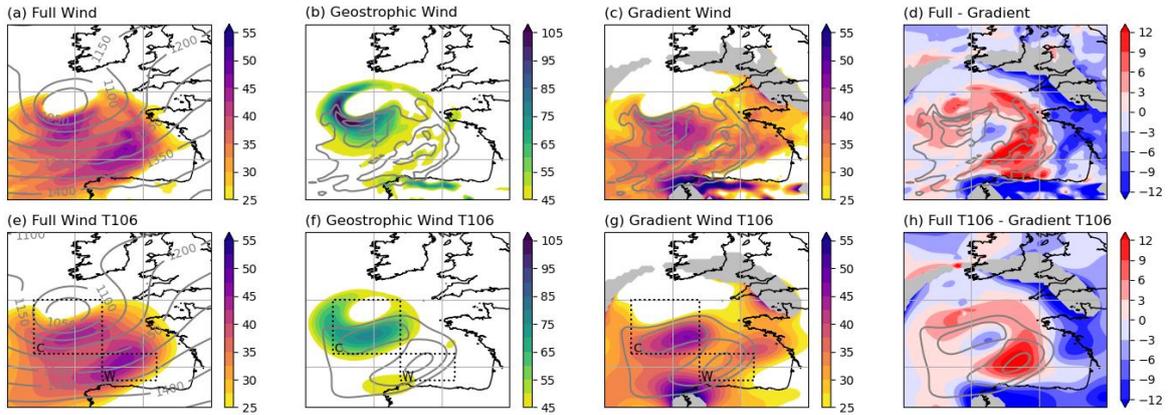

*Figure S3: As Figure S2 but for IFS HRES initialised at 00 UTC 1 November 2023.*

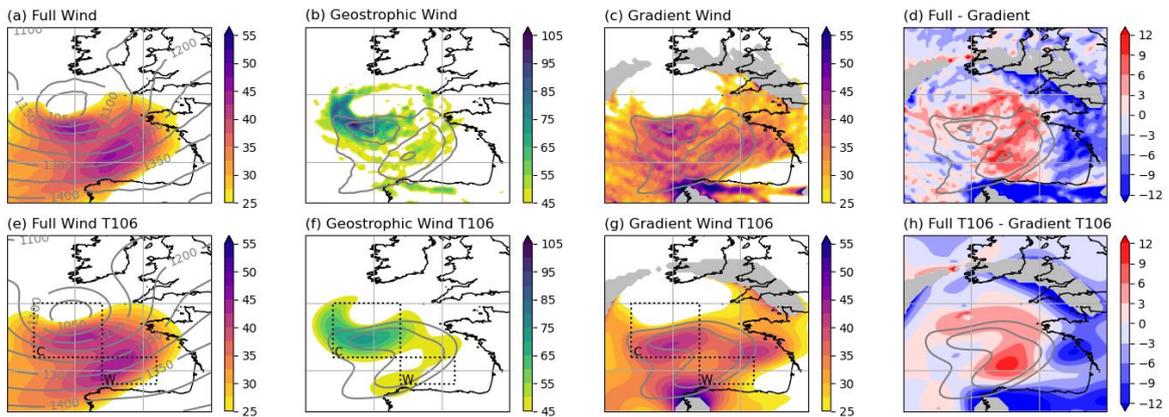

*Figure S4: As Figure S2 but for GraphCast.*

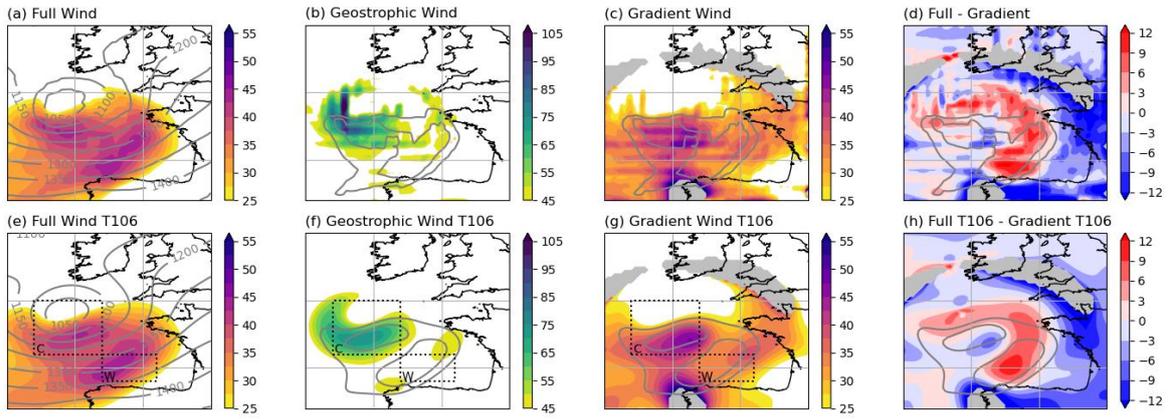

*Figure S5: As Figure S2 but for PanguWeather.*

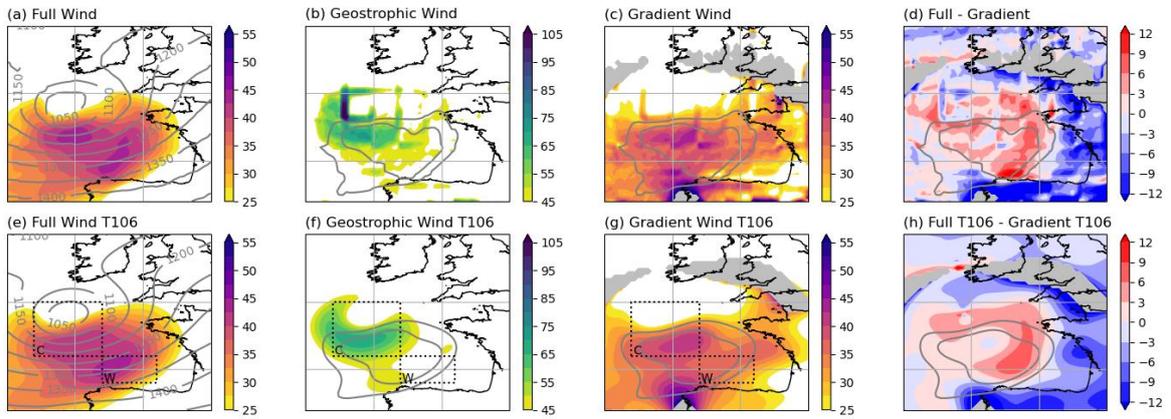

*Figure S6: As Figure S2 but for FourCastNet.*

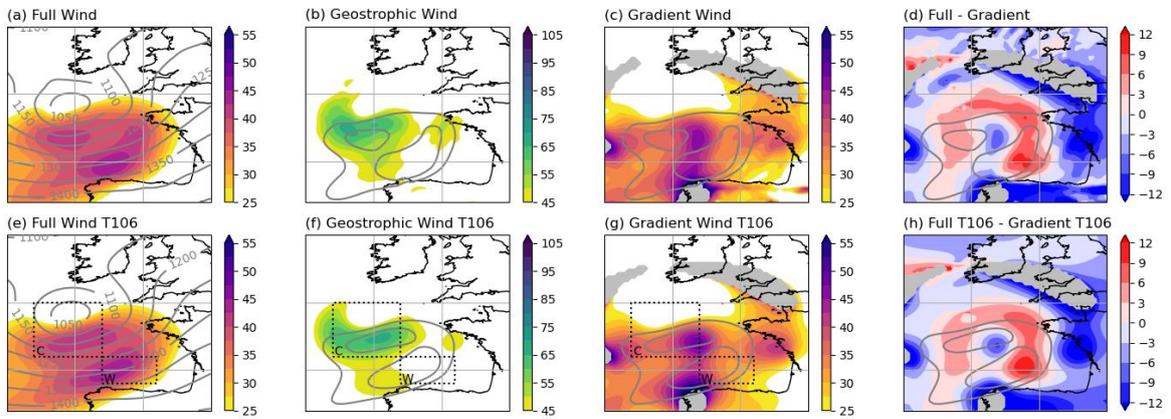

*Figure S7: As Figure S2 but for FourCastNet v2.*

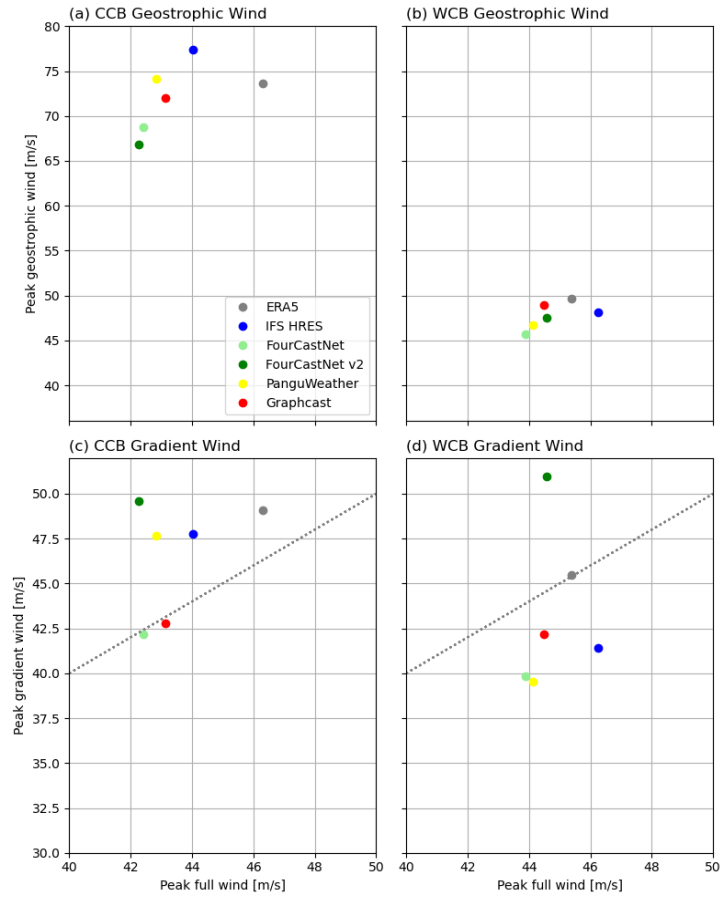

*Figure S8: A summary of the relationship between peak full winds and peak geostrophic (a and b) and peak gradient (c and d) winds in the cold (a and c) and warm (b and d) conveyor belt regions of Storm Ciarán. Each dot represents the maximum wind in the boxes indicated in Figures S1-S6 using the T106 smoothed fields.*